\documentclass[runningheads]{llncs}
\usepackage{graphicx}
\usepackage{mathtools} 
\usepackage[ruled,linesnumbered,vlined]{algorithm2e}
\usepackage{subcaption} 
\usepackage{booktabs}
\usepackage{todonotes}
\usepackage{amssymb}
\renewcommand{\epsilon}{\varepsilon}

\newcommand{\sobol}{Sobol\kern-0.15em\'{ } }

\begin{document}
%
\title{Algorithm Selection with Probing Trajectories:\\ Benchmarking the Choice of Classifier Model}
\titlerunning{Algorithm Selection with Probing-Trajectories}
%
\author{Quentin Renau\inst{1}\orcidID{0000-0002-2487-981X} \and
Emma Hart\inst{1}\orcidID{0000-0002-5405-4413}}

\authorrunning{Q. Renau and E. Hart}

\institute{Edinburgh Napier University, Edinburgh, Scotland, UK 
\email{\{q.renau,e.hart\}@napier.ac.uk}}

\maketitle              
\begin{abstract}
Recent approaches to training algorithm selectors in the black-box optimisation domain have advocated for the use of training data that is `algorithm-centric' in order to encapsulate information about how an algorithm performs on an instance, rather than relying on information  derived from features of the instance itself. \textit{Probing 
trajectories} that consist of a sequence of objective performance per function evaluation obtained from a short run of an algorithm have recently shown particular promise in training accurate selectors. However, training models on this type of data requires an appropriately chosen classifier given the sequential nature of the data. There are currently no clear guidelines for choosing the most appropriate classifier for algorithm selection using time-series data from the plethora of  models available. To address this, we conduct a large benchmark study using $17$ different classifiers and three types of trajectory on a classification task using the BBOB benchmark suite using both leave-one-instance out and leave-one-problem out cross-validation. In contrast to previous studies using tabular data, we find that the choice of classifier has a significant impact, showing that \textit{feature-based} and \textit{interval-based} models are the best choices.

\end{abstract}

\section{Introduction}
\label{sec:intro}
Per-instance algorithm selection (AS) that uses a machine-learning (ML) model to choose the most appropriate solver from a portfolio has shown much promise in both continuous and combinatorial optimisation domains~\cite{kerschke_automated_2019}. Designing a selector has two important facets: determining what kind of input will be used to the selector, and the choice of the ML model itself. 

With regard to the former, many approaches rely on an input vector that captures either features derived from the description of instance~\cite{AlissaSH23} or features derived from the landscape induced by the instance and an objective function, e.g. Exploratory Landscape Features (ELA)~\cite{mersmann_exploratory_2011}.  These methods can be considered as \textit{instance-centric} as they are independent of the choice of solver. On the other hand, recent approaches to  AS have taken an \textit{algorithm-centric} approach, in which input to a model is derived from the performance of an algorithm on an instance: for example, Renau {\em et. al.}~\cite{RenauH24,RenauH24-2} use  time-series data as input to a classifier, in what is termed a trajectory-based approach. The trajectory contains a short sequence of the objective values obtained at each evaluation of a solution over a short time period. 

The choice of ML model used in the selector is dependent on the type of input data.  Feature-based approaches to AS have often used tree-based classifiers such as Random Forests~\cite{JankovicVKNED22}, and there is a growing trend of exploiting new deep-learning architectures based on transformers, e.g. ~\cite{gorishniy2021revisiting}.  However, a different type of classifier is required to cope with time-series data when using trajectory-based approaches. In~\cite{RenauH24}, a time-series forest classifier was used~\cite{TSF} as the model, but alternative approaches were not evaluated.

Kostovska {\em et. al.}~\cite{kostovska2023comparing} assessed the influence of the ML model used in AS on single-objective black-box problems in the context of feature-based classifiers. They find that `the choice of model has only a minor impact on the AS performance, as long as it is a method that demonstrates good performance on tabular data in general settings'.   However, given that time-series data is fundamentally different to tabular data in that it is an ordered sequence of values and that specialised time-series classifiers fall into a wide range of categories, it is unclear whether the result from \cite{kostovska2023comparing} holds when considering trajectory-based AS.

\textbf{Contribution}: In this paper, we address the question raised above, i.e. "\textit{To what extent does the choice of ML model used in algorithm selection matter when using trajectory-based data?}". We conduct an extensive evaluation of $17$ time-series-based classifiers using the BBOB test suite as a benchmark, with a portfolio of three solvers. Experiments are conducted both in the Leave-one-instance-out (LOIO) and Leave-one-problem-out (LOPO) settings, where the latter is known to be harder~\cite{DerbelLVAT19}. We find that in contrast to tabular data, the choice of ML model in the time-series setting has a significant impact on the accuracy of the model. 
We improve the gain over ELA features obtained in~\cite{RenauH24} from $3\%$ to $7\%$ for similar budgets and we show that we can use even lower budgets of function evaluations and still obtain a $2\%$ gain over ELA features.

Additionally, in the LOPO setting, we show that some functions are extremely difficult to classify correctly regardless of the model chosen, while a small set of four models obtain an accuracy of $\ge 90\%$ on $11$ out of $24$ functions.

The outline of this paper is as follows.
Section~\ref{sec:related} gives an overview of the background and related work.
Section~\ref{sec:method} describes the data used, the methods for obtaining probing-trajectories, describes the models benchmarked and experiments conducted in this paper.
Section~\ref{sec:results} describes the results obtained with the probing-trajectories on an algorithm selection task.
Section~\ref{sec:discussion} provides insights into the results and describes some limitations of our work.
Finally, Section~\ref{sec:conclusion} highlights concluding remarks and future work.

\section{Background and Related Work}
\label{sec:related}

The majority of previous work in algorithm selection is performed using information describing an \textit{instance} as input to a selector. In the continuous optimisation domain, it is common to use Exploratory Landscape Analysis (ELA)~\cite{mersmann_exploratory_2011} to extract features used as training data~\cite{RenauDDD21,JankovicD20}, while in combinatorial optimisation there has been a recent trend towards feature-free approaches that use information that directly describes an instance: in the bin-packing domain, the sizes of each item to be packed are directly used as input~\cite{AlissaSH23}, while in the TSP domain, images that directly indicate city locations have been used~\cite{seiler2020deep}.
However, these approaches all take an \textit{instance-centric} view of a problem: that is, the input to a selector describes instance data only and is independent of the execution of any algorithm. Intuitively, incorporating some measure of algorithm performance into the input to a selector would  seem beneficial.

Recognising this, recent work has begun to address this, using information derived from running a solver as input to a selector. For example, Jankovic \textit{et. al.}~\cite{JankovicED21} propose  using ELA features extracted from the search trajectory of an algorithm as training data. Their approach gave encouraging results but was outperformed by classical ELA features computed on the full search space. In~\cite{JankovicVKNED22},  features obtained from algorithm trajectories are successfully used to  determine whether to switch an algorithm during the course of solving an instance. In~\cite{CenikjPDKE23},  standard statistics (mean, standard deviation, minimum, maximum) are extracted from function evaluations at each generation and used as input to a classifier that predicts which of the $24$ BBOB function the trajectory belongs to. Renau {\em et. al.}  directly use algorithm trajectories to train a classifier to predict the best of three solvers using the BBOB test-suite, showing that these approaches outperform classifiers trained on ELA features, and later show that trajectories can also be used to train a classifier to detect whether an instance is easy or hard for a portfolio of solvers in~\cite{RenauH24-3}. Although not specifically concerned  with algorithm selection, the work of~\cite{PitraRH19} is also worthy of mention in taking an algorithm perspective by utilising information incorporated in CMA-ES \textit{state variables} to train a surrogate model to predict performance.

Regardless of the type of input used in an algorithm selector, it is important to  consider what type of machine learning (ML)  model is best suited to the task. Kostovska {\em et. al.}~\cite{kostovska2023comparing} study the impact of the choice of ML model when classifying tabular data, evaluating four different ML models (tree-based and deep-learning-based) on three AS approaches: regression, classification, and pair-wise classification.  They find while per-instance algorithm selection has impressive potential, the ML technique is of minor importance.

Given that time-series data obtained from algorithm trajectories is significantly different to tabular data and requires a different family of ML model to be used as a classifier, we conduct a similar benchmarking exercise to that described in~\cite{kostovska2023comparing} to understand the influence of the choice of ML models when training an algorithm selector on trajectories.

\section{Methods}
\label{sec:method}
We benchmark a suite of $17$ different ML models that use time-series data as input to a classifier to better understand what type of classifier facilitates algorithm selection. The methods for obtaining the trajectory data and the models selected for evaluation are described below. 

\subsection{Trajectories}
\label{sec:traj}
We use the same data as described in \cite{RenauH24} to benchmark $17$ classifiers on the noiseless Black-Box Optimisation Benchmark (BBOB) from the COCO platform~\cite{cocoJournal} using data from three algorithms: CMA-ES~\cite{HansenO01}, Particle Swarm Optimisation (PSO)~\cite{KennedyPSO95}, and Differential Evolution (DE)~\cite{StornP97}. As in \cite{RenauH24}, all data is obtained from \cite{dataDiederick}. 

Input to each classifier is a \textit{probing-trajectory}. This consists of a time-series consisting of the first $n$ function evaluations from a run of an algorithm. Three types of trajectories are defined.

\begin{itemize}
    \item \textit{Best}: At each function evaluation, the best objective value seen so far is recorded per algorithm;
    \item \textit{Current}: the objective value obtained after every evaluation is recorded per algorithm;
    \item \textit{All}: the trajectories from each algorithm are concatenated to form a single trajectory (for either 'Best' or 'Current'), for example, in this paper ALL best refers to the concatenation of the \textit{Best} trajectories from CMA-ES, DE, and PSO.
\end{itemize}

For a single instance, four trajectories can therefore be obtained per instance (one from each algorithm and one that concatenates the individual trajectories).  
Three of these trajectories contain data from one run of an algorithm; the final trajectory contains data from one run of each of three algorithms. 
Each trajectory is treated as a time-series and is used as input to an algorithm selector.
The length of this time-series depends on the number of concatenated algorithms and the number of generations used by each algorithm.
We match the experiments performed in~\cite{RenauH24} by testing trajectories in which the number of generations  $g \in \{2,7\}$.  We use data from $5$ instances of each of the $24$ BBOB functions in dimension $10$, and perform $5$ runs per instance. 
Therefore, we obtain a dataset of $24\times5\times5=600$ trajectories.

\subsection{Algorithm Selection}
\label{sec:classif}
In order to replicate the experiments in~\cite{RenauH24}, we consider an algorithm selection task  as a classification task, i.e., given a time-series representing an instance, the output is the algorithm to use on that particular instance.
The winning algorithm per instance is defined as the algorithm having the best median target value after $100{,}000$ function evaluations and is used as the label for the classifier.
No single algorithm outperforms the others on all $24$ functions: CMA-ES is the best performing algorithm for $11$ functions, DE for $7$, and PSO for $6$.

\paragraph{Classification Models}
Probing-trajectories consist of a time-series and thus require the use of a specialised time-series classifier .
In~\cite{RenauH24}, the authors used a Rotation Forests~\cite{RodriguezKA06} from the \emph{sktime} package~\cite{sktime}\footnote{version 0.33.0}. 
In other work using trajectory-based input to a classifier~\cite{RenauH24-3} a Long Short Term Memory (LSTM) network~\cite{lstm} was used: these networks have been shown to deal well with sequential information.
In this paper, our goal is to benchmark time-series classifiers to understand how the choice of classifier influences performance.
As such, we evaluated $17$ time-series classifiers, $16$ from the \emph{sktime} package and the LSTM from~\cite{RenauH24-3}.
We excluded other classifiers that gave errors on our data or had an inference computation time that would exceed the running time of an optimisation algorithm.
The list of classifiers is composed of $3$ Deep Learning (DL) models, $3$ distance-based (D) models, $2$ feature-based (F) models, $3$ interval-based (I) models, $3$ kernel-based (K) models, $1$ shapelet-based (S) model, $1$ sklearn-based (sk) model, and $1$ dummy model.
Default parameters are initially used for all mentioned classifiers.
The classifiers are defined as follows:
\begin{itemize}
    \item{\bf LSTM (DL)} as defined in~\cite{RenauH24-3}. Code for this model can be found at~\cite{dataHardness};
    \item{\bf Time Convolutional Neural Network (CNN) (DL)} as defined in~\cite{CNN};
    \item{\bf Multivariate Time Series Transformer for Classification} (MVTST) (DL)] as defined in~\cite{MVTST};
    \item{\bf K-nearest neighbors (D)} adapted version of the scikit-learn package~\cite{scikit-learn} for time-series data;
    \item{\bf Proximity Stump (D)} models a decision stump, i.e., a one-level decision tree, which uses distance to partition the data;
    \item{\bf ShapeDTW (D)} as defined in~\cite{ShapeDTW};
    \item{\bf Catch22 (F)}  as defined in~\cite{catch22};
    \item{\bf Summary (F)} extracts statistics on the data and builds a Random Forest~\cite{Breiman01} classifier;
    \item{\bf Random Interval Spectral Ensemble (I)} as defined in~\cite{RISE};
    \item{\bf Supervised Time Series Forest (I)} as defined in~\cite{STSF};
    \item{\bf Time Series Forest Classifier (I)} as defined in~\cite{TSF};
    \item{\bf Support Vector Classifier (K)} adapted version of the scikit-learn~\cite{scikit-learn} SVC for time series data;
    \item{\bf Arsenal (K)} as defined in~\cite{Arsenal};
    \item{\bf Rocket (K)} as defined in~\cite{Rocket};
    \item{\bf Shapelet Transform Classifier (S)} as defined in~\cite{Shapelet};
    \item{\bf Rotation Forests (sk)} as defined in~\cite{RodriguezKA06};
    \item{\bf Dummy} which simply outputs the most represented class in the training data.
\end{itemize}

\paragraph{Automated Configuration of Models}
We perform automated configuration of the parameters of the best models.
If not stated otherwise, we use \emph{irace}~\cite{irace}\footnote{version 3.5} to tune the models parameters.
We use \emph{irace}'s default parameters.
The number of evaluations used is $5{,}000$ for cheap models to evaluate and $1{,}000$ for more expensive models.
As CMA-ES has the smallest population size, $10$, we choose to tune models using the best trajectory of CMA-ES for $2$ generations.
This choice is motivated by the computation time of models, i.e., shorter time-series minimises the training and inference times of models.
We then transfer the results found for CMA-ES best trajectory to all other trajectories.

\paragraph{Validation Procedure}
We perform two types of validations.
As performed in~\cite{KostovskaJVNWED22}, we perform a \emph{leave-one-instance-out (LOIO) cross-validation} and we compute \emph{the overall accuracy}.
We train classifiers using runs from all $24$ functions on all except one instance.
Data from the left out instance is used as the validation set.
Overall, $24\times4\times5 = 480$ inputs are used to train the model while the remaining $24\times1\times5 = 120$ inputs are used for validation.

As performed in~\cite{DerbelLVAT19}, we also perform a \emph{leave-one-problem-out (LOPO) cross-validation} and we compute \emph{the overall accuracy}.
We train classifiers using runs from $23$ functions on all instances and data from the left out function is used as validation set.
Overall, $23\times5\times5 = 575$ inputs are used to train the model while the remaining $1\times5\times5 = 25$ inputs are used for validation.

\section{Results}
\label{sec:results}
In this section, we first present the results obtained with default models on a LOIO cross-validation (Section~\ref{sec:default_loio}), followed by results obtained with the tuning of models on a LOIO cross-validation (Section~\ref{sec:tuning}).  
Finally, we present results obtained on the LOPO cross-validation (Section~\ref{sec:default_lopo}).
We discuss the results in Section~\ref{sec:discussion}.

\subsection{Default Models on LOIO Cross-Validation}
\label{sec:default_loio}
In this section, we compare $17$ default time-series classifiers on the algorithm selection task presented in Section~\ref{sec:classif}.
We use a LOIO cross-validation setting, i.e., models are trained on all functions for four instances and tested on the remaining instance.
The $17$ models are trained for best and current trajectories for all four trajectories described in Section~\ref{sec:traj}, resulting in a total of $136$ classifiers.
As similar tendencies can be observed for all trajectories used as input, we will only present results for PSO trajectories for $2$ generations and CMA-ES trajectories for $7$ generations.
We provide results for the other trajectories in the supplementary materials~\cite{dataBenchTS}.

Figure~\ref{fig:loio} compares the classification accuracy of the classifiers trained using PSO trajectories for $2$ generations (Figure~\ref{fig:loio_pso}) and CMA-ES trajectories for $7$ generations (Figure~\ref{fig:loio_cma}).
We can observe  the same general pattern in Figure~\ref{fig:loio}.
Classifiers that perform well when trained using PSO trajectories also perform well when trained with CMA-ES trajectories. This observation can be extended to all other studied trajectories.

Renau {\em et. al.}~\cite{RenauH24} observed that training a classifier using the \textit{best} trajectory is better than training using the \textit{current} trajectory if running for a small number of generations, and vice versa for a large number of generations.
However this result is not seen in our data --- it only holds for $2$ of the $17$ models: \textit{Rotation Forest} and \textit{MVTSTransformer}.
Moreover, we observe that the Rotation Forest classifier used in~\cite{RenauH24} is never the best performing model for any of the trajectories studied. 

Furthermore, for both CMA-ES trajectories and PSO trajectories, we observe that some models perform on par or even worse than the Dummy classifier (recall this simply outputs the dominant class in the training data): this applies to $5$ models trained using PSO trajectories and $3$ for the CMA-ES trajectories. Two of these poor-performing models are kernel-based (\textit{Arsenal} and \textit{Rocket}), two are Deep Learning-based (\textit{CNN} and\textit{ MVTSTransformer}) and one is distance-based (\textit{ProximityStump}).
On the contrary, all feature-based and interval-based models perform amongst the best performing classifiers.
These two categories of classifier are the only ones in which all models from the category obtain  performances above the Dummy classifier.
Other models such as \textit{ShapeletTransform} or \textit{TimeSeriesSVC} display average performance.
They outperform the Dummy classifier but are outperformed by at least $10\%$ median accuracy by the best performing models.

\begin{figure}
\centering
\begin{subfigure}{.45\textwidth}
  \centering
  \includegraphics[width=\linewidth]{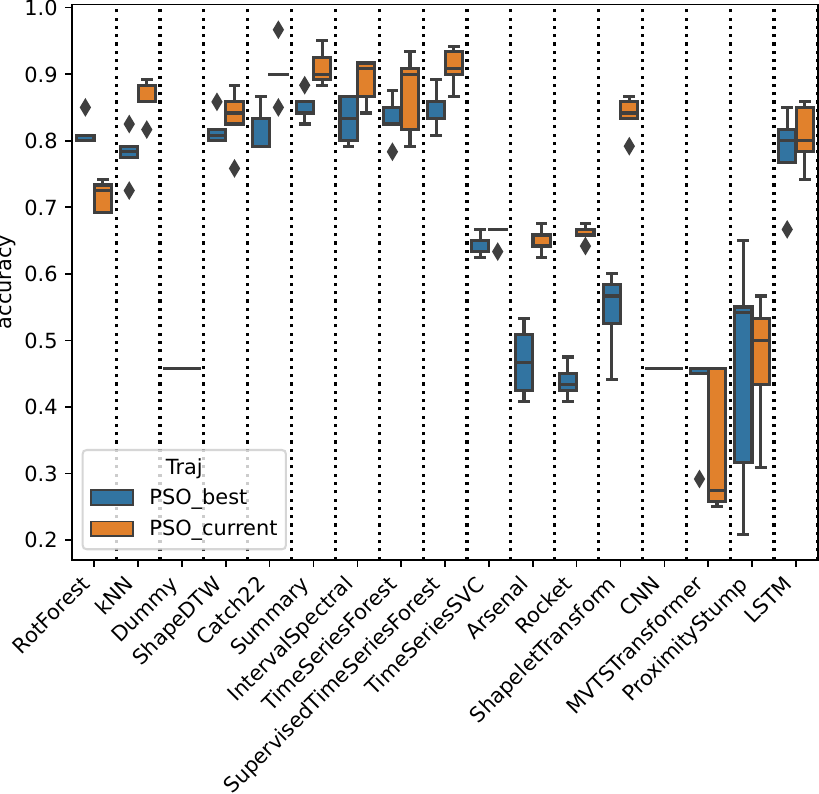}
  \caption{PSO $2$ generations}
  \label{fig:loio_pso}
\end{subfigure}
\begin{subfigure}{.45\textwidth}
  \centering
  \includegraphics[width=\linewidth]{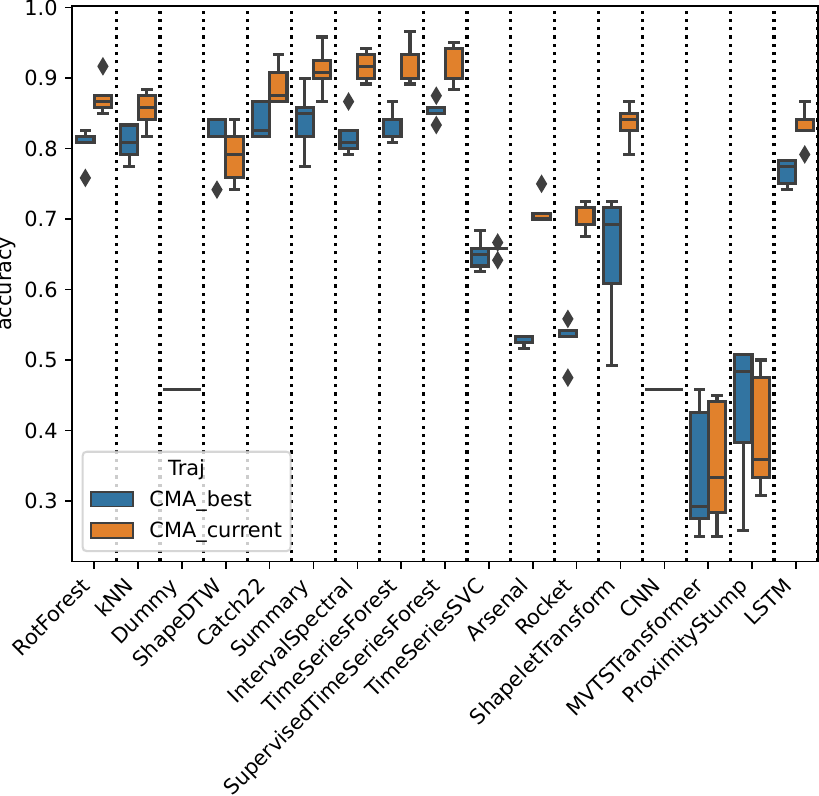}
  \caption{CMA-ES $7$ generations}
  \label{fig:loio_cma}
 \end{subfigure}
 \caption{Accuracy of classification on the LOIO cross-validation for best-so-far and current probing-trajectories for $2$ generations (PSO) and $7$ generations (CMA-ES).}
 \label{fig:loio}
\end{figure}

\subsection{Tuning of Models on LOIO Cross-Validation}
\label{sec:tuning}
In order to improve the performance of models, we configure the parameters of the best performing models seen in Figure~\ref{fig:loio}.
From these results, we identify $9$ candidate models to tune.
After removing models that are too expensive to tune and models where altering some parameters  caused errors in the Python package, we are left with $6$ models to compare. These models and the parameters that can be tuned are described below:
\begin{itemize}
    \item \textbf{kNN}: number of neighbours (between $1$ and $30$), weight function (uniform or distance), and distance measure between time series (ten possibilities);
    \item \textbf{ShapeDTW}\footnote{\label{note} $1{,}000$ evaluations used by \emph{irace}}: number of neighbors (between $1$ and $30$) and the descriptor function (six possibilities);
    \item \textbf{Time Series Forest}: number of estimators (between $10$ and $500$) and the minimum length of an interval (between $3$ and $30$);
    \item \textbf{Summary}: statistics functions used ($8191$ possible combinations) and the quantiles to compute (eight possibilities); 
    \item \textbf{Rotation Forest}\footref{note}: number of estimators (between $10$ and $500$), minimum and maximum size of an attribute group (between $3$ and $30$), and the proportion of cases to be removed per group;
    \item \textbf{Supervised Time Series Forest}: number of estimators (between $10$ and $500$). The tuning of this model was performed using a grid search dividing the search space in $50$ values.
\end{itemize}

As mentioned in Section~\ref{sec:method}, tuning is  performed using the \textit{best} trajectory from runs of CMA-ES.  The tuned parameters are then transferred to all trajectories.

We find that all \textit{Supervised Time Series Forest} configurations have similar performances and thus we discard this classifier from the rest of the study on the configuration of parameters.
Distributions of accuracies for all parameters tested overlap.
We performed Kolmogorov-Smirnov tests on pairs of distributions and we cannot reject any null hypothesis, i.e.,  the two samples tested come from the same distribution (see supplementary material~\cite{dataBenchTS} for the plot). 
For all other models,  we find that tuning improves performance. Details of the tuned configurations can be found in Table~\ref{tab:params}.

\begin{table}[h]
\caption{Tuned parameters for the six models.}
\label{tab:params}
\centering
\begin{tabular}{|c|c|}
\hline
Models             & Parameters                                                                                                                  \\ \hline
kNN                & $4$ neighbors, uniform weights, twe distance                                                                                  \\
ShapeDTW           & $4$ neighbors, raw descriptor                                                                                                 \\
Time Series Forest & $460$ estimators, $3$ minimum interval                                                                                          \\
Summary            & \begin{tabular}[c]{@{}c@{}}mean,min, max, kurtosis, variance, \\ nb unique and count statistics, $0.25$ quantile\end{tabular} \\
Rotation Forest    & \begin{tabular}[c]{@{}c@{}}$367$ estimators, min group of $10$,\\  max group of $19$, remove proportion of $0.2364$\end{tabular}    \\ \hline
\end{tabular}
\end{table}

\begin{figure}
\centering
\begin{subfigure}{.45\textwidth}
  \centering
  \includegraphics[width=\linewidth]{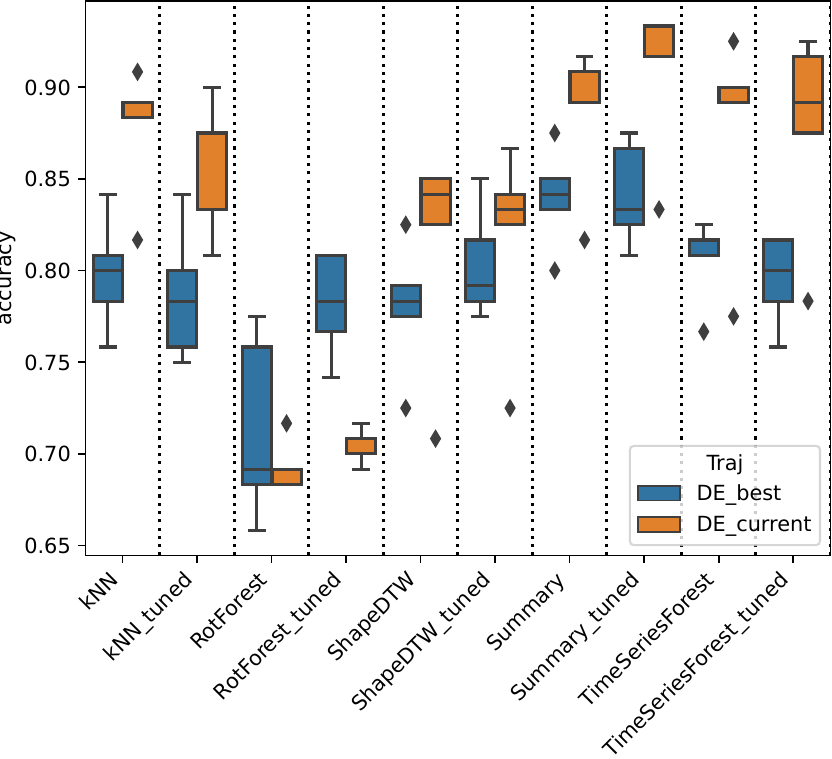}
  \caption{DE $2$ generations}
  \label{fig:loio_de}
\end{subfigure}
\begin{subfigure}{.45\textwidth}
  \centering
  \includegraphics[width=\linewidth]{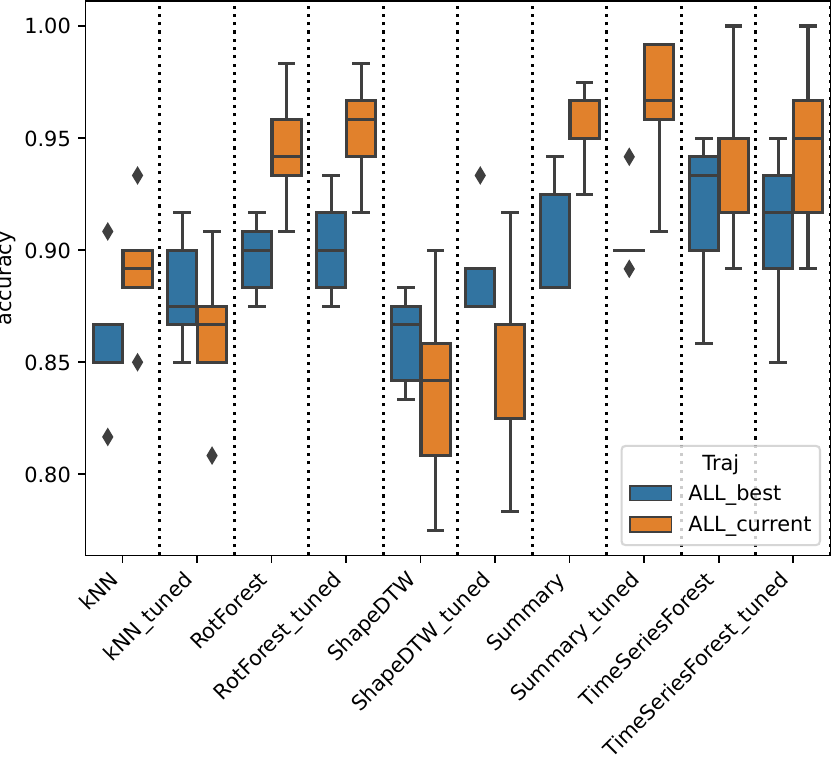}
  \caption{ALL $7$ generations}
  \label{fig:loio_all}
 \end{subfigure}
 \caption{Accuracy of classification on the LOIO cross-validation for best-so-far and current probing-trajectories for $2$ generations (DE) and $7$ generations (ALL) for default and tuned models.}
 \label{fig:loio_tuning}
\end{figure}

Figure~\ref{fig:loio_tuning} compares the results obtained from tuned/default configurations of the $5$ models described above for models trained using (a) DE trajectories and (b) the concatenated ALL trajectories, i.e., we transfer the parameters found when tuning for CMA-ES best trajectory to DE and ALL trajectories.
Results are provided using both best and current trajectories in each case.
We observe that transferring the parameters found for models tuned with CMA-ES best trajectories improves the accuracy of classification in most settings, i.e., only kNN trained on the DE best trajectory (Figure~\ref{fig:loio_de}),  kNN on the ALL current, and Time Series Forest on the ALL best trajectory (Figure~\ref{fig:loio_all}) do not improve in performance when compared to the default parameters.

The configured \textit{Summary} classifier on current trajectories is the best performing model  in Figure~\ref{fig:loio_tuning}.
This result holds for models trained on $6$ of the $8$ types of trajectories used to train the models in this paper, i.e. from CMA-ES, DE, PSO, and ALL trajectories, calculated over $2$ or $7$ generations. 
Using trajectories from  CMA-ES at both $2$ and $7$ generations,  we observe that the tuned \textit{Summary} classifier is outperformed by the tuned \textit{Time Series Forest} classifier but obtains the rank of second best in terms of performance.

\subsection{Default Models on LOPO Cross-Validation}
\label{sec:default_lopo}
In this section, we consider a LOPO cross-validation.
Models are trained on all but one function and validated on  instances from the remaining function.
Given the construction of BBOB, this task is much harder than a LOIO cross-validation as functions are purposely designed to be diverse.

\begin{figure}
\centering
\includegraphics[width=0.8\linewidth]{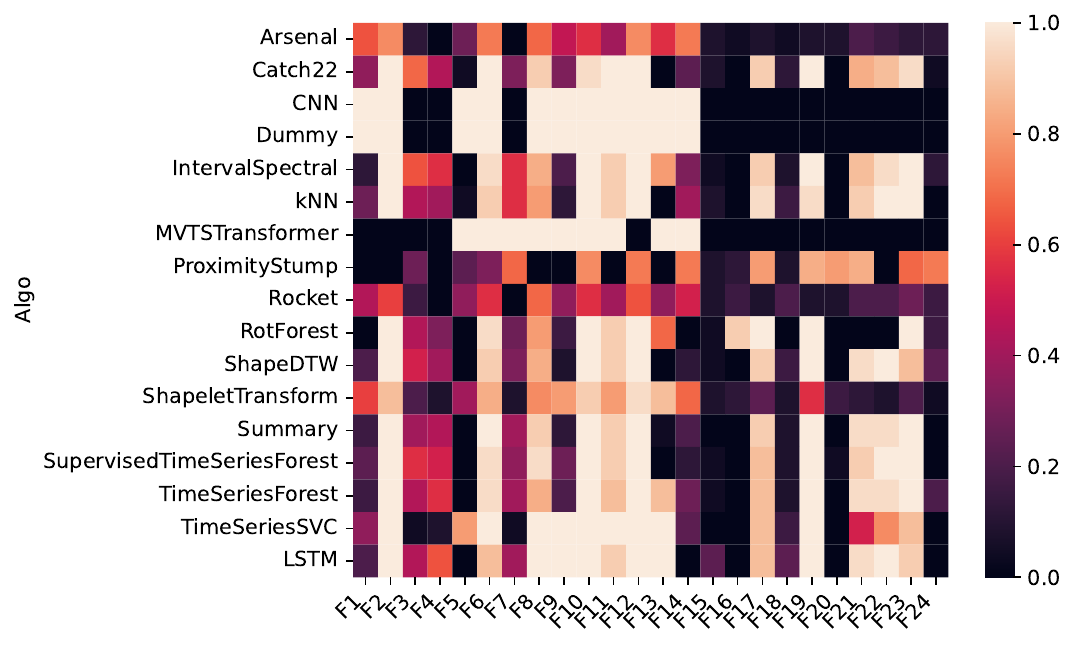}
\caption{Heatmap of classification accuracy on the LOPO cross-validation for CMA-ES best-so-far probing-trajectories for $2$ generations for default. $x-$axis represents the functions left out for validation.}
\label{fig:lopo}
\end{figure}

Figure~\ref{fig:lopo} displays the results for all models on the LOPO cross-validation where the $x-$axis represents the function left out for validation, e.g., column F24 represents a training set composed of functions F1 to F23 and a validation set composed of F24. The trajectory used to train the models in each case is the CMA-ES best trajectory over $2$ generations.

\paragraph{Insights into model performance}
As expected, the performance obtained in the LOPO setting  is poorer than  using LOIO cross-validation, with a maximum average accuracy of $61.3\%$ across all functions obtained by the LSTM model, which is the best model on average across all functions.

\begin{figure}
\centering
\includegraphics[width=0.6\linewidth]{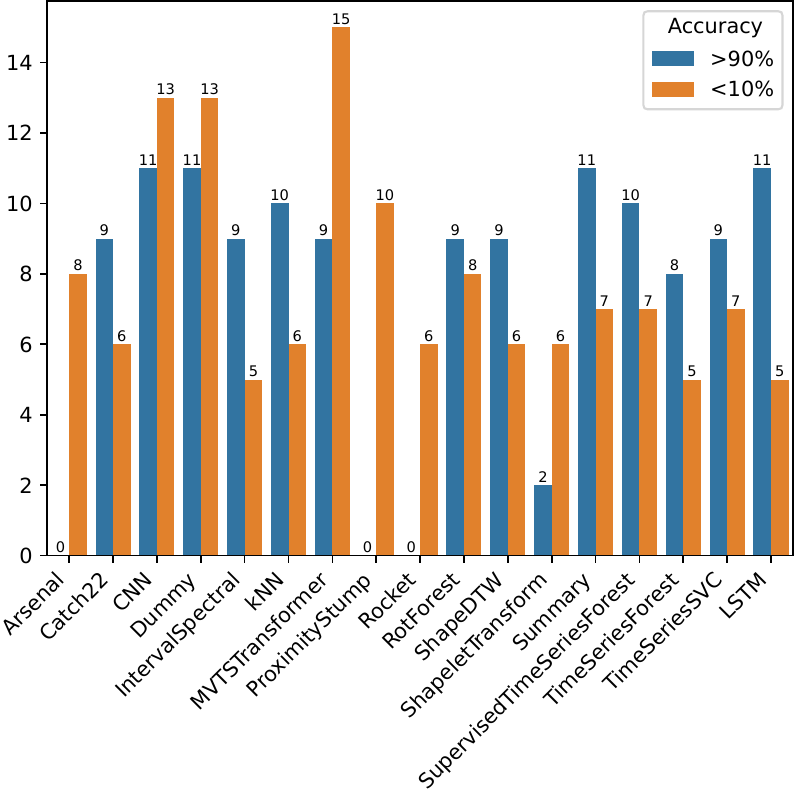}
\caption{Number of functions with accuracy above $90\%$ or below $10\%$ for each model.}
\label{fig:bar_model}
\end{figure}

Figure~\ref{fig:bar_model} displays the number of functions where the accuracy of a model is below $10\%$ or above $90\%$.
Four models obtain an accuracy greater or equal to $90\%$ on $11$ functions: \textit{CNN, Dummy, LSTM}, and \textit{Summary}. The high performance of \textit{Dummy} can be explained by the frequency of winning algorithms in the training data, i.e., CMA-ES wins on $11$ functions and Dummy outputs the majority class in training, which is therefore CMA-ES. Looking at the behaviour of \textit{CNN}, we observe that the model learns the same behaviour as \textit{Dummy}, i.e., it always outputs CMA-ES as an answer. Thus, the only two models that learn to predict the correct algorithm rather than simply outputting the majority class are \textit{LSTM} and \textit{Summary}, obtaining at least $90\%$ on $11$ functions.

The lowest number of functions where models obtain an accuracy lower than $10\%$ is $5$.
Three models obtain an accuracy lower than $10\%$ on $5$ functions: \textit{LSTM}, \textit{Time Series Forest} and \textit{Random Interval Spectral Ensemble} 
These three models also achieve at least $90\%$ on $11$, $8$ and $9$ functions respectively indicating that these three models can perform well and rarely perform poorly.
Along with \textit{LSTM}, the other best performing model, \textit{Summary}, obtains a lower than $10\%$ accuracy on $7$ functions indicating that it does rarely perform poorly, even though it does not match the best three models in that matter.
Overall, the best performing model on a LOPO cross-validation is \textit{LSTM} followed by \textit{Summary} for the number of well predicted functions.

\paragraph{Insights into the difficulty of predicting functions.}
We observe that for some functions, it is relatively straightforward to train a high-performing selector, while for others, it is very difficult.

\begin{figure}
\centering
\includegraphics[width=0.6\linewidth]{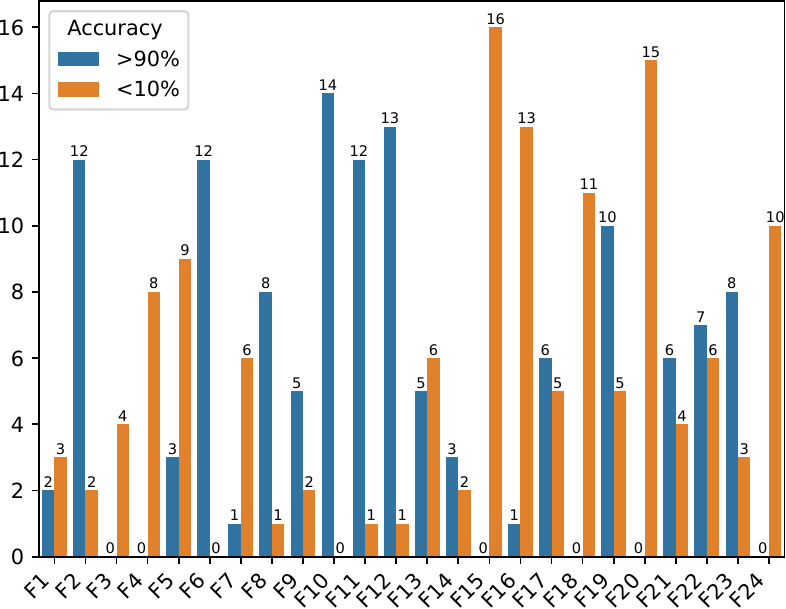}
\caption{Number of models  with accuracy above $90\%$ or below $10\%$ for each function.}
\label{fig:bar_function}
\end{figure}

Figure~\ref{fig:bar_function} displays the number of models where the accuracy on a function is below $10\%$ or above $90\%$.
Well predicted functions are F10, F12, F2, and F6 with $14$, $13$, $12$, and $12$ models obtaining more than $90\%$ accuracy on them respectively.
No model performs below $56\%$ on F10 and $32\%$ on F6; only one model performs below $10\%$ on F12 while two perform below this value on F2.

On the contrary, the most difficult functions to perform LOPO on are F15, F20, and F16 with $16$ models, $14$ models, and $12$ models respectively performing below $10\%$ accuracy.
All models perform poorly on F15 with a maximum median accuracy of $24\%$ (LSTM). 
On F20, the  best median accuracy obtained is $80\%$ from one model (\textit{Proximity Stump}).
Only one model performs above $90\%$ accuracy on F16: \textit{Rotation Forest}.

\section{Discussion}
\label{sec:discussion}
In Section~\ref{sec:results} we showed results obtained from training different models of algorithm selectors using two cross-validation strategies (LOIO,LOPO) using sixteen different  types of training data as input: i.e. the best and current trajectories obtained from three algorithms,  the ALL trajectories obtained by concatenating three best or three current trajectories, with trajectories collected over both  $2$ or $7$ generations.
From these experiments, we make a general observation that models performances in terms of accuracy are not linked to the type of cross-validation used and generally is invariant to the type of trajectory, i.e., the best performing model for one trajectory is likely to perform well on another trajectory.
\textit{Kernel-based} and \textit{deep learning-based} classifiers (except LSTM) consistently perform poorly on the algorithm selection task. In fact, their performance is often comparable to those of the \textit{Dummy} classifier which always predicts the majority class. Hence, we suggest that these types of models may not be suited for algorithm selection using time-series data.
Concerning \textit{deep learning} models, it is well-known in the machine-learning literature that they often require a lot of fine tuning to perform well, in terms of both the architecture and setting weights~\cite{BaratchiWLRHBO24}.
In our experiments, we only used off-the-shelf architectures for these models, hence fine tuning the architectures of these models may improve their performance.

We observe that \textit{feature-based} and \textit{interval-based} models consistently perform well.
One feature-based model (\textit{Summary}) and one interval-based model (\textit{Time Series Forest}) are often ranked as the best performing model in an experiment.
Hence, we recommend that one of these two models is used as a default when training algorithm selectors on raw time-series data.

\textit{Limitations:} The study used a single test suite (BBOB). Although this is used extensively in the black-box optimisation community~\cite{RenauH24-3,KostovskaJVNWED22}, it is known that it does not generalise well~\cite{vermetten23a}, especially in the LOIO cross-validation setting. We compensate for this by also performing a LOPO cross-validation, but extending the study to other test suites would be beneficial to understand better whether the results we obtained generalise across other benchmark suites, in both continuous and combinatorial optimisation. In addition, our tuning study only using CMA-ES trajectory data in tuning model parameters (Section~\ref{sec:tuning}). Although the tuned model clearly transfers well to models that use trajectories from the other algorithms as input, tuning a model on the specific trajectory type of interest might further improve results.

\section{Conclusion}
\label{sec:conclusion}

This article addressed the question of the extent to which the choice of machine-learning model used in the context of algorithm selection with time-series input influences the accuracy of the learned classifier. A large study benchmarked $17$ classifiers from multiple families of time-series classifiers using the BBOB test suite, in both LOIO and LOPO settings.

Overall, we observe that classifiers that perform well in the LOIO setting also perform well in the
LOPO setting. Additionally, the best performing classifiers for short trajectories of $2$ generations are also the best performing classifiers for longer trajectories composed of $7$ generations and this for all considered trajectory types. Two classes of models are consistently ranked as the best performing models for all settings considered: \textit{Summary} and \textit{Time Series Forest}. These models perform similarly despite the fact the  classifiers operate in a very different manner.
\textit{Summary} is a feature-based model that extracts statistics on the time-series and builds a Random Forest classifier on these statistics while \textit{Time Series Forest} is an interval-based model building an ensemble of decision trees on random intervals of the time-series. Hence we recommend the use of one of these two models to perform algorithm-selection using time-series input. We also performed an automated configuration of model parameters using a single trajectory type, and then transfer the tuned configurations to models trained on different trajectories. We show that tuned version of the models outperform the default configurations even when parameters learned on one type of trajectory are directly transferred to models that are trained on trajectories. Nevertheless, we expect that  tuning the models using the trajectory type that will eventually be used to perform algorithm selection could bring further improvements.

In Renau {\em  et al.}~\cite{RenauH24}, the  authors used a LOIO setting using a model trained using a  \textit{Rotation Forest} classifier.
With this setting, the authors obtained a $3\%$  gain in accuracy over a classifier trained using ELA features and  using a similar budget ($560$ and $500$ function evaluations).
By tuning this classifier using automated algorithm configuration, we show that we can further increase the gain in accuracy to $5\%$. Interestingly, in a low budget setting using a model trained on the  CMA-ES current trajectory with $70$ function evaluations and using tuned \textit{Time Series Forest} classifiers, we also obtain a $2\%$ gain over ELA features but using more than $7$ times fewer function evaluations. However, using the tuned \textit{Summary} classifier, we further increase  the gain to $6\%$ using a similar function evaluation budget.

Obvious next steps include extending the benchmarking approach to other datasets both in continuous and combinatorial domains. Moreover, in this paper, we perform algorithm-selection from a classification perspective. As it is also common to train regressors as algorithm-selectors,   benchmarking regressors for time-series inputs should also be undertaken in future.

\begin{credits}
\subsubsection{\ackname} Authors are supported by funding from EPSRC award number: EP/V026534/1.

\clearpage
\subsubsection{\discintname}
The authors have no competing interests to declare that are
relevant to the content of this article. 
\end{credits}


\bibliographystyle{splncs04}
\bibliography{references}

\end{document}